\documentclass[nohyperref]{article}

\usepackage{microtype}
\usepackage{graphicx}
\usepackage{subfigure}
\usepackage{booktabs} % for professional tables

\usepackage{hyperref}

\usepackage[accepted]{wfvml2022}

\usepackage{amsmath}
\usepackage{amssymb}
\usepackage{mathtools}
\usepackage{amsthm}

\usepackage[capitalize,noabbrev]{cleveref}

\theoremstyle{plain}

\theoremstyle{definition}

\theoremstyle{remark}

\usepackage[textsize=tiny]{todonotes}

\usepackage{bm}
\usepackage{dsfont} 
\usepackage{wrapfig}
\usepackage{cancel}
\usepackage{graphicx}

\wfvmltitlerunning{Correcting Model Bias with Sparse Implicit Processes}

\begin{document}

\twocolumn[
\wfvmltitle{Correcting Model Bias with Sparse Implicit Processes}

% unless you've provided the [accepted] option to the wfvml2022



\begin{wfvmlauthorlist}
\wfvmlauthor{Simón Rodríguez Santana}{icmat}
\wfvmlauthor{Luis A. Ortega}{uam}
\wfvmlauthor{Daniel Hernández-Lobato}{uam}
\wfvmlauthor{Bryan Zaldivar}{uv}
\end{wfvmlauthorlist}

\wfvmlaffiliation{icmat}{Institute of Mathematical Sciences (ICMAT-CSIC), Madrid, Spain.}
\wfvmlaffiliation{uam}{Escuela Polit\'ecnica Superior, Universidad Aut\'onoma de Madrid, Spain.}
\wfvmlaffiliation{uv}{Institute of Corpuscular Physics, University of Valencia and CSIC, Spain.}

\wfvmlcorrespondingauthor{Simón Rodríguez Santana}{simon.rodriguez@icmat.es}

\wfvmlkeywords{Bayesian Machine Learning, Approximate Inference, Stochastic Processes, Implicit Processes, Model Bias}

\vskip 0.3in
]


% The \wfvmlEqualContribution command is standard text for equal contribution.

\printAffiliationsAndNotice{}  % leave blank if no need to mention equal contribution
% \printAffiliationsAndNotice{\wfvmlEqualContribution} % otherwise use the standard text.

\begin{abstract}
Model selection in machine learning (ML) is a crucial part of the Bayesian learning procedure. Model choice may impose strong biases on the resulting predictions, which can hinder the performance of methods such as Bayesian neural networks and neural samplers. 
On the other hand, newly proposed approaches for Bayesian ML exploit features of approximate inference in function space with implicit stochastic processes (a generalization of Gaussian processes). 
The approach of \textit{Sparse Implicit Processes} (SIP) is particularly successful in this regard, since it is fully trainable and achieves flexible predictions. 
Here, we expand on the original experiments to show that SIP is capable of correcting model bias when the data generating mechanism differs strongly from the one implied by the model. 
We use synthetic datasets to show that SIP is capable of providing predictive distributions that reflect the data better than the exact predictions of the initial, but wrongly assumed model. 
\end{abstract}

%%%%%%%%%%%%%%%%%%%%%%%%%%%%%%%%%%%%%%%%%%%%%%%%%%%%%%%%%%%%%%%%%%%%%%%%%%%%%%%
%%%%%%%%%%%%%%%%%%%%%%%%%%%%%%%%%%%%%%%%%%%%%%%%%%%%%%%%%%%%%%%%%%%%%%%%%%%%%%%
% MAIN BODY
%%%%%%%%%%%%%%%%%%%%%%%%%%%%%%%%%%%%%%%%%%%%%%%%%%%%%%%%%%%%%%%%%%%%%%%%%%%%%%%
%%%%%%%%%%%%%%%%%%%%%%%%%%%%%%%%%%%%%%%%%%%%%%%%%%%%%%%%%%%%%%%%%%%%%%%%%%%%%%%

\section{Introduction}
\label{sec:introduction}

Approximate Bayesian inference techniques extend the usual point-wise predictions of already successful methods, \emph{e.g.} neural networks (NNs) \cite{blei2017variational}, to full predictive distributions. These techniques are seen as fast, general and approachable ways for obtaining more complete information about the predictions obtained \cite{blundell2015weight, gal2016uncertainty, gal2016dropout}. The estimated uncertainty of the predictions can account for different sources of uncertainty present in the modelling process. Moreover, this extra information can play a crucial role in sensitive real-world scenarios, where decisions may be taken depending on the certainty of the predictions \cite{mcallister2017concrete}. However, these estimates will always be subject to the selection of the proper model for the data and the task at hand. 
% Therefore, it is natural to look for general-purposed methods inside this framework that may encompass very different behaviors, not neglecting key features that may be potentially important for the data \cite{knoblauch2019generalized}. 

In Bayesian modelling, one uses probability distributions to assign degrees of belief to the parameter values of these models \cite{graves2011practical, BlundellCKW15}. Nonetheless, Bayesian learning presents several intrinsic difficulties in this context, \emph{e.g.} the choice of meaningful priors when using complex systems such as Bayesian NNs (BNNs), as well as dealing with the intractability of the calculations required. Usually, these issues are avoided by resorting to approximations of the original posterior distribution. However, this also induces certain biases that can have potentially decisive effects on the resulting predictive distribution. Moreover, the popularity of bigger models further complicates the learning process due to the numerous symmetries and strong dependencies that arise in parameter-space \cite{sun2019functional, coker2021wide}. These factors may compromise the performance of regular approximation techniques.

Function-space inference has become a prominent approach to solve some of the intrinsic issues of approximate inference in parameter-space. Performing inference in function-space simplifies the issues related to the complexity of the optimization space. In particular, approaches based on \textit{implicit stochastic processes} (IPs) have gathered a lot of attention \cite{ma2019variational,sun2019functional}. Until recently, the most advanced methods were unable to simultaneously yield flexible predictive distributions and a fully trainable framework capable of tuning all of its parameters. The first method to do both things while remaining based on IPs is \emph{Sparse Implicit Processes} (SIP) \cite{santana2022sparse}. Additionally, and maybe surprisingly, SIP's predictive distribution seems capable of correcting the model bias imposed by the selection of a given prior. In general, if the selected prior does not suit the data at hand, the exact predictive distribution can become quite different from the data distribution. Here we show that SIP is capable of providing predictive distributions that closely resemble the data, even if the underlying prior model may not be the best choice. We test out this feature by comparing SIP with other function-based models. We show that, due to the changes in the approximate inference objective function, SIP's predictive distribution is much more robust to the model choice than other approaches. This leads to better generalization properties, capturing complex patterns in the predictive distribution. 

\section{Background}
\label{sec:background}

Before we explore the model bias correction in SIP, let us briefly introduce the key ideas behind this framework.

\paragraph{Approximate inference in parameter space}
\label{sec:approximate_inference_in_parameter_space}

Consider some data $\mathcal{D}=\{\mathbf{x}_i,y_i\}_{i=1}^N$, a prior distribution $p(\mathbf{w})$ over the model parameters $\mathbf{w}$ and a likelihood model $p(\mathbf{y}|\mathbf{w},\mathbf{X})$. Let distribution $q_\phi(\mathbf{w})$ approximate the exact posterior $p(\mathbf{w}|\mathbf{y},\mathbf{X})$ resulting from the Bayes rule. In variational inference, $q_\phi(\mathbf{w})$ is obtained maximizing the evidence lower bound (ELBO) \cite{jordan1999introduction}:
\begin{align}
\mathcal{L}(\phi) &= 
\mathds{E}_{q_\phi(\mathbf{w})}[\log p(\mathbf{y}|\mathbf{w},\mathbf{X})] - \text{KL}(q_\phi(\mathbf{w})|p(\mathbf{w}))
        \label{eq:lower_bound_vi}
\end{align}
where $\text{KL}(\cdot, \cdot)$ represents the Kullback-Leibler divergence between $q$ and $p$. Usually, $q$ is assumed to be parametric for complex models such as Bayesian NNs (BNNs), imposing also independence among the components of $\mathbf{w}$ \cite{BlundellCKW15, graves2011practical}.
%  This is equivalent to minimizing $\text{KL}(q_\phi(\mathbf{w})|p(\mathbf{w}|\mathbf{y},\mathbf{X}))$. 

An implicit model can be used for $q$ \emph{i.e.}, $q_\phi(\mathbf{w}) = \int q_\phi(\mathbf{w}|\bm{\epsilon}) p(\bm{\epsilon}) d \bm{\epsilon}$, with $\bm{\epsilon}$ some random noise \cite{mescheder2017adversarial,santana2020adversarial}. If $\bm{\epsilon}$ is high-dimensional and the model defining $q_\phi(\mathbf{w}|\bm{\epsilon})$ is expressive enough, $q_\phi(\mathbf{w})$ can emulate almost any function. However, since $q$ lacks a closed-form density, the KL term becomes intractable. A solution here is rewriting it as the result of an auxiliary classification problem. The optimal value of this auxiliary problem is precisely $ T_{\omega^\star} (\mathbf{w}) = \log q_\phi(\mathbf{w}) - \log p(\mathbf{w})$ \cite{mescheder2017adversarial}.

Finally, instead of minimizing the regular KL-divergence between $q$ and 
the posterior, other approaches employ the more general $\alpha$-divergences, which
include the KL-divergence as a particular case \cite{hernandez2016black}. This has 
shown to overall improve the final results \cite{ma2019variational, santana2020adversarial, wenzel20a}

\paragraph{Inference with Implicit Processes}
\label{par:implicit_processes}

Implicit processes (IPs) can be seen as a general framework over different stochastic processes, \emph{e.g.} Gaussian processes can be understood as a particular case of IPs. An IP is defined as a collection of random variables $f(\cdot)$ such that the joint distribution of any finite set of evaluations $(f(\mathbf{x}_1), \cdots , f(\mathbf{x}_N))^\text{T}$ is determined by the generative process:
\begin{align}
\mathbf{z} \sim p(\mathbf{z})\,,& \quad  f(\mathbf{x}_i) = g_\theta (\mathbf{x}_i, \mathbf{z}), \quad
        \forall \mathbf{x}_i \in \mathcal{X},
\end{align}
where $\mathbf{z}$ is some random variable that summarizes the randomness, $\theta$ represents the parameters of the process and $\mathcal{X}$ is the input space \cite{ma2019variational}. We define $f(\cdot) \sim \mathcal{IP}(g_\theta(\cdot, \cdot), p_{\mathbf{z}})$ to indicate that $f$ is sampled from the corresponding IP with parameters $\theta$, using samples from $p(\mathbf{z})$ (denoted as $p_{\mathbf{z}}$). This definition of IPs enables many models to be described with the same language, \emph{e.g.} BNNs, warped GPs, neural samplers (NS), and deep GPs, among others \cite{snelson2004warped,ma2019variational,damianou2013deep}. 

Conducting approximate inference with IPs has become a promising research topic with increasing interest \cite{ma2019variational, sun2019functional}. However, the complexity of dealing with IPs leads to important issues in the formulation of such methods. The approach in \citet{ma2019variational} relies on approximating the marginal likelihood of the prior IP by the marginal likelihood of a GP. This leads to Gaussian predictions, which may lack flexibility in complex situations. The same issue arises in \citet{ma2021}. On the other hand, the method in \citet{sun2019functional} is not capable of updating the prior model parameters according to the data due to the usage of a spectral gradient estimator. This can result in important performance losses, since specifying a meaningful prior in complex models such as BNNs is a non-trivial task \cite{knoblauch2019generalized}.

\paragraph{Sparse Implicit Processes (SIP)}
\label{par:sparse_implicit_processes}

In \citet{santana2022sparse}, SIP is described as being the first general-purposed IP-based method that remains fully trainable and with flexible predictive distributions. This is achieved through an inducing point approximation in the likes of sparse GP approximations \cite{titsias2009variational}. Scalability is therefore ensured by conducting inference with $M \ll N$ number of inducing points. Denoting $\overline{\mathbf{X}}$ as the set of inducing points, the IP values at these input locations is $\mathbf{u}=(f(\overline{\mathbf{x}}_1),\ldots,f(\overline{\mathbf{x}}_M))^\text{T}$. The exact posterior  $p(\mathbf{f},\mathbf{u}|\mathcal{D})$ is then approximated by 
\begin{align}
\label{eq:posterior_approximation}
    q(\mathbf{f}, \mathbf{u}) &= p_\theta(\mathbf{f}| \mathbf{u}) q_\phi(\mathbf{u})\,,
\end{align}
where $\phi$ are the parameters of the implicit distribution $q_\phi(\mathbf{u})$, and $\theta$ are the parameters of the IP prior. Crucially, $p_\theta(\mathbf{f}| \mathbf{u})$ is approximated by a Monte Carlo GP approximation as in \citet{ma2019variational}. The resulting functional-ELBO is:
\begin{align}
\label{eq:fELBO_SIP}
        \mathcal{L}(\phi,\theta) = \mathds{E}_{q_{\phi}}[ \log p(\mathbf{y}|\mathbf{f}) ] - \text{KL}(q_\phi(\mathbf{u})|p_\theta(\mathbf{u}))\,.
\end{align}
Similar to \eqref{eq:lower_bound_vi}, the first term can be estimated by MC sampling, but the KL term lacks closed-form solution. To estimate the exact log-ratio, an auxiliary classification problem is used, making $ \text{KL}(q_\phi(\mathbf{u})|p_\theta(\mathbf{u})) = \mathds{E}_q \left[ T_{\omega^\star}(\mathbf{u}) \right]$, being $T_{\omega^\star}(\mathbf{u})$ the classifier's optimal value \cite{mescheder2017adversarial}. 

SIP is capable of tuning its prior parameters by using the symmetrized KL-divergence \cite{santana2022sparse}. This leads to a better fit of the prior to data, even though the final objective is no longer a lower bound on the original objective, as in \citet{ma2019variational}. Moreover, SIP resorts to $\alpha$-divergence energy function, present in other methods \cite{hernandez2016black,ma2019variational,santana2020adversarial}, leaving as final objective 
\begin{equation}
\begin{aligned}
\label{eq:alpha_f-elbo}
    \mathcal{L}_\alpha^\star(\phi,\theta) & =  \textstyle \frac{1}{\alpha} \sum_{i=1}^N \log \mathds{E}_{q_{\phi,\theta}} 
	[p(y_i|f_i)^\alpha ] \\
	& \quad -  \textstyle \frac{1}{2} \left[ \text{KL}(q_\phi | p_\theta) 
	+ \text{KL}(p_\theta | q_\phi) \right]\,.
\end{aligned}
\end{equation}

Finally, $p_\theta(\mathbf{f}|\mathbf{u})$ in \eqref{eq:posterior_approximation} is approximated by the conditional of a GP with the same covariance and mean function as the prior IP, as in \citet{ma2019variational}, where the mean and covariances are empirically estimated. Predictions in new locations $\mathbf{x}^\star$ are given via Monte Carlo using $S$ samples. Using $\mathbf{u}_s \sim q_\phi(\mathbf{u})$, the predictions are
\begin{align}
\label{eq:predictions_SIP}
        p(f(\mathbf{x}_\star)|\mathbf{y},\mathbf{X}) \approx  \textstyle \frac{1}{S} \sum_{s=1}^S p_\theta(f(\mathbf{x}_\star)|
        \mathbf{u}_s)\,.
\end{align}
The resulting distribution is a mixture of Gaussians. Using multiple samples, this enables SIP to reproduce flexible predictive distributions that need not be Gaussian, as can be seen in the original experiments \cite{santana2022sparse}.

In a general sense, the first term in the r.h.s. of \eqref{eq:alpha_f-elbo} forces predictions to explain the observed data, while the second term makes predictions similar to the prior in regions where no data points are available. Here, the approximations required in SIP can be understood in a similar fashion to the recent work in \citet{knoblauch2019generalized}. It is the combination of the posterior approximation and the approximate objective function which allows SIP to produce flexible predictions. Moreover, and perhaps surprisingly, SIP seems capable of correcting the imposed model bias by the selection of a prior not suited to the data at hand. The original experiments in \citet{santana2022sparse} hinted at this: the functions generated using the BNN priors are NN samples contaminated with Gaussian noise, which are normally distributed. This means that, through exact Bayesian inference, using such prior models one should not obtain bimodal or heteroscedastic predictive distributions. This is shown in \citet{santana2022sparse} through the results of HMC on synthetic data. On the other hand, SIP seems capable of \textit{correcting} this model bias, providing more flexible predictive distributions than the ones obtained through exact Bayesian inference. We explore this feature in the following experiments. 

\begin{table*}[h]
\small
\caption{Results for training the system in $20$ different instances of the bimodal and heterocedastic datasets.}
\vspace*{0.1cm}
\label{tab:results}
\begin{center}
\begin{tabular}{llcclcc}
\multicolumn{1}{c}{} &  & \multicolumn{2}{c}{Bimodal data} &  & \multicolumn{2}{c}{Heteroscedastic data} \\ \cline{3-4} \cline{6-7} 
\\ [-1em]
\multicolumn{1}{c}{} &  & Exact GP & SIP &  & Exact GP & SIP \\ \cline{3-4} \cline{6-7} 
RMSE &  & \textbf{4.988 $\pm$ 0.003} & 4.994 $\pm$ 0.004 &  & \textbf{1.340 $\pm$ 0.002} & 1.342 $\pm$ 0.002 \\
NLL &  & 3.026 $\pm$ 0.001 & \textbf{2.093 $\pm$ 0.003} &  & 1.714 $\pm$ 0.001 & \textbf{1.449 $\pm$ 0.001} \\
CRPS &  & 2.860 $\pm$ 0.089 & \textbf{2.429 $\pm$ 0.003} &  & 0.727 $\pm$ 0.033 & \textbf{0.678 $\pm$ 0.001}
\end{tabular}
\end{center}
\end{table*}

\begin{figure*}[th]
\vspace*{-0.6cm}
	\begin{tabular}{cc}
    \includegraphics[width=0.48\textwidth]{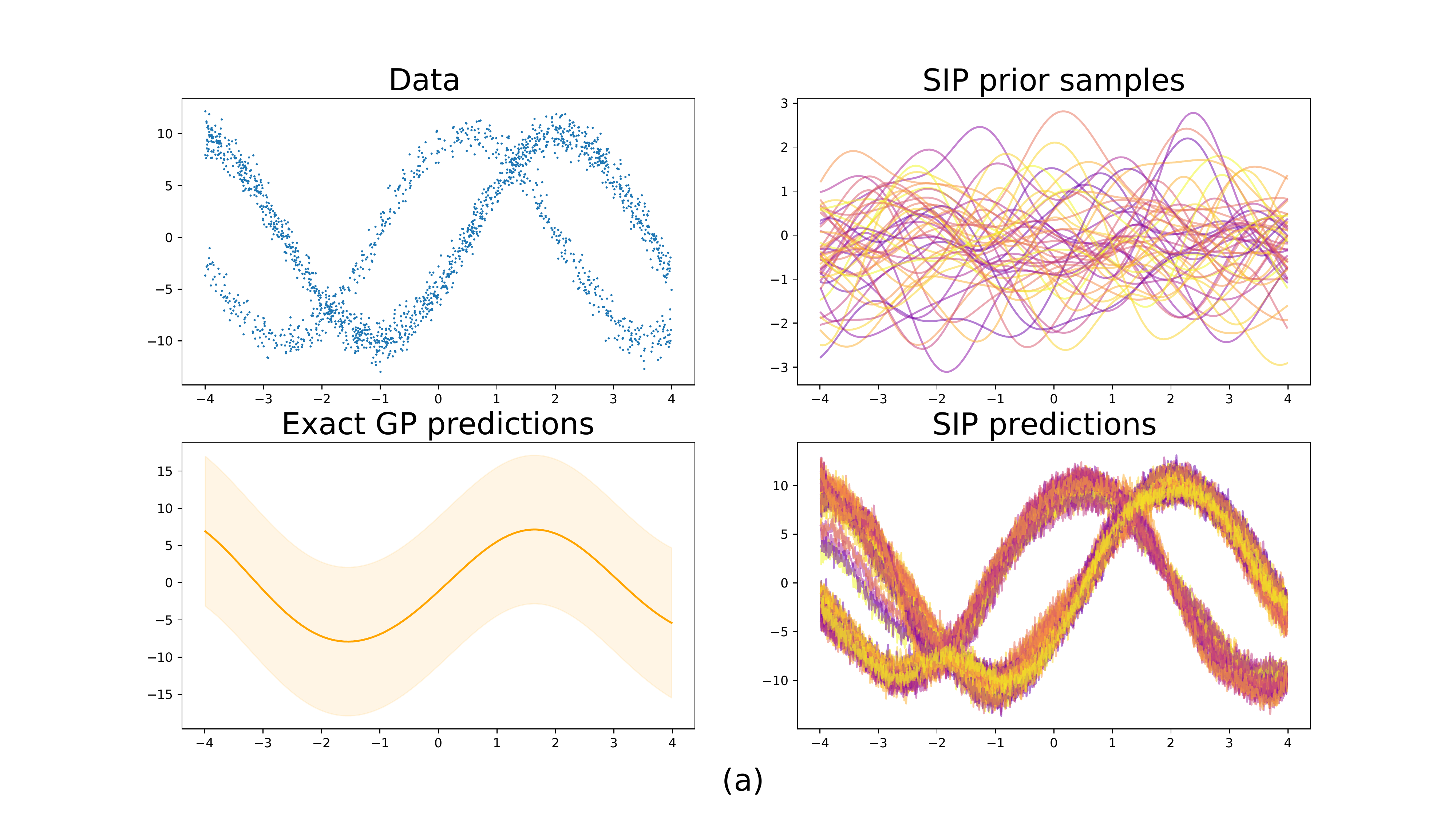}
    \includegraphics[width=0.48\textwidth]{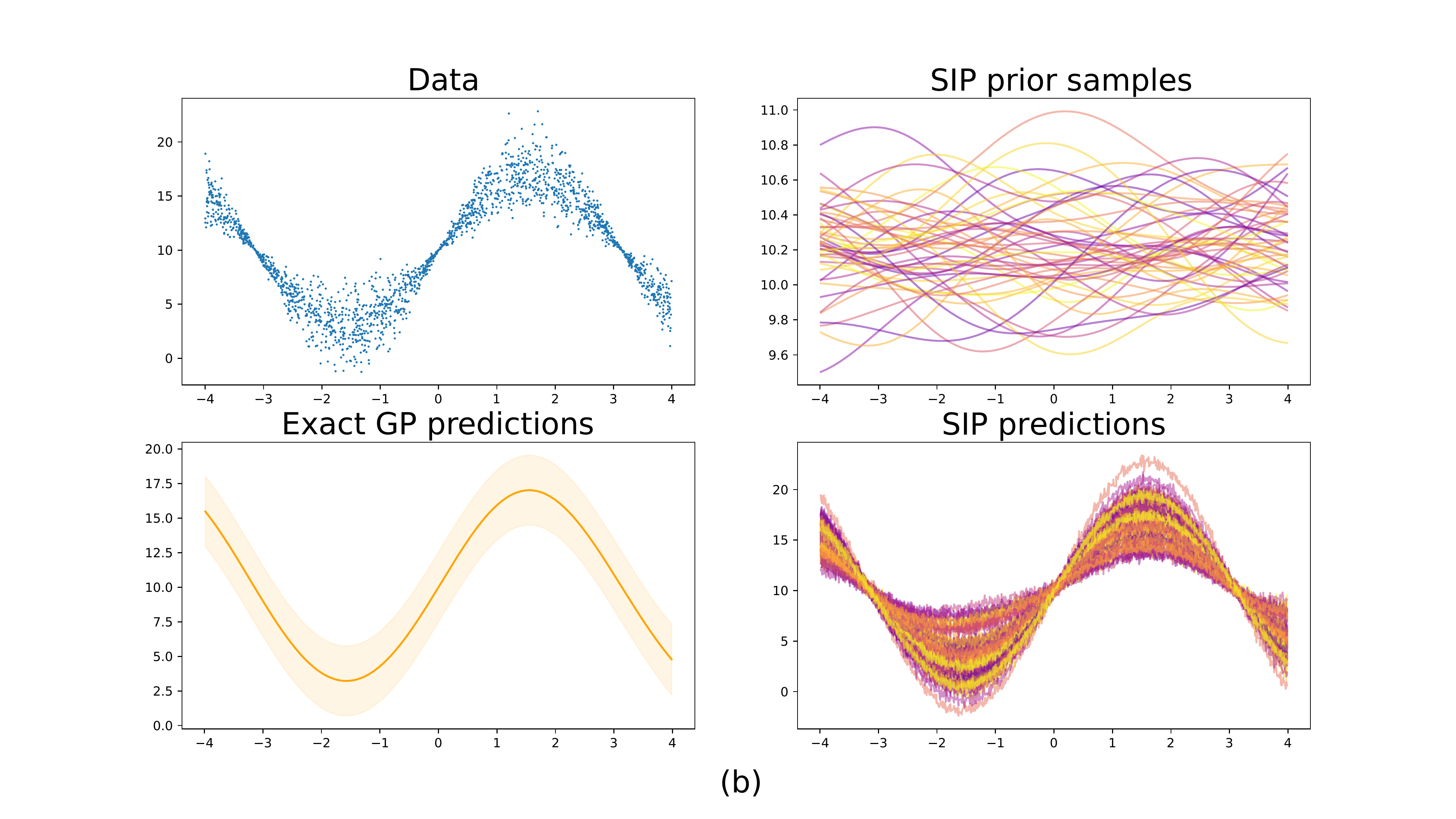}
    \end{tabular}
    \vspace*{-0.5cm}
    \caption{Results for the experiments comparing an exact GP model and SIP with the same GP prior with bimodal data \textbf{(a)} and heterocedastic data \textbf{(b)}. For both experiments we show, clockwise and beginning from the top-left: the original data (\textit{blue}), samples from the GP prior of SIP, samples from SIP's predictive distribution, and the exact GP predictive distribution (\textit{orange}). Best seen in color.}
    \label{fig:results}
\end{figure*}

\section{Experiments}
\label{sec:experiments}

In the following tests we compare SIP's predictions with those of regular GPs. We study the behavior of SIP when forcing certain model biases in the implementation. By employing a GP prior, the associated posterior distribution will be that of a GP, which is normally distributed and can be obtained in an exact manner. Therefore, the GP predictions will serve as ground truth for the expected behavior of the exact posterior distribution. Using a GP prior in SIP, the exact GP posterior distribution is also available. However, if SIP maintains the behavior shown by \citet{santana2022sparse}, we expect it to avoid this model-formulation bias and output non-Gaussian predictions if the data presented is also clearly non-Gaussian. 
For this comparison, the exact GP model makes use of a white and a RBF kernel using the  \href{https://scikit-learn.org/stable/modules/generated/sklearn.gaussian_process.GaussianProcessRegressor.html}{GP regression module} in \texttt{sklearn}. In SIP we also use a GP prior implemented in the form of a wide 1-layer BNN with 500 units and cosine activation functions, as described in \citet{rahimi2007random}. The weights are initialized using a standard Gaussian, while the biases are sampled from $\mathcal{U}[0, 2\pi]$ (only for the hidden layer). The kernel length and amplitude parameters are initialized to 1. Finally, SIP's posterior IP is parametrized using a 2-layer BNN with 50 units per layer, leaky-ReLU activations and 50 inducing points. 

Following the original results in \citet{santana2022sparse}, we expect SIP to compensate for this model bias in the formulation and output flexible predictions. To show this, we employ both an exact GP model and SIP in the same two synthetic cases present in the original article, one with bimodal data and another with heteroscedastic data. To measure the performance of each method in both datasets, we generate the data 20 different times following the same equations with different random seeds. We train both methods in each generated dataset and measure the performance metrics to average across the 20 different instances.

\paragraph{Bimodal dataset}
To generate the bimodal data we first sample 1000 values for $x$ from $\mathcal{U}(-4, 4)$. Then, we obtain one of the two possible $y$ values with probability $0.5$. Fixing $\epsilon \sim \mathcal{N}(0,1)$, these two values are:
\begin{align}
    y_1 = 10 \cos(x - 0.5) + \epsilon, \:\:\:\:\:\:\:\:
    y_2 = 10 \sin(x - 0.5) + \epsilon,
    \nonumber
\end{align}

In Fig.\ref{fig:results}a we plot the the bimodal data in the top-left figure (in blue) and the resulting predictions for both SIP and the exact GP model. In the bottom-left we have the predictions of the exact GP model, which, as expected, follow a Gaussian behavior centered in the mean of the original data. The shaded orange regions represent two standard deviations from the GP predictive mean, marked here by the dark orange line in the middle. On the other hand, in the right-side figures we see SIP's results. In the top figure we plot samples from the GP prior used for SIP (the same as for the exact GP). Note that these samples do not follow the data in the same fashion of the original article since they are simply samples from the trained prior GP, although the associated hyperparameters are fitted to the data. Finally, in the bottom-right we have samples from the predictive distribution of SIP. This distribution is clearly bimodal, which differs strongly from the predictive distribution of the exact GP, which would be the result from the process of exact inference. SIP's predictions follow closely the original data, achieving a more representative predictive distribution that is not available with exact inference from the initial GP prior imposed. This model bias is, in practice, being corrected by the flexibility of the approximate inference method in SIP, which follows the original behavior observed in  \citet{santana2022sparse}. Interestingly, each function sampled from SIP is located alongside one mode. It is through the multiple samples where this flexible behavior is obtained.

\paragraph{Heteroscedastic dataset}
To obtain the heteroscedastic data, we again sample $1000$ values for $x$ from $\mathcal{U}(-4, 4)$ and then obtain the respective $y$ value as 
\begin{equation}
    y = 7\sin (x) + \epsilon \sin(x) + 10, \:\:\:\:\: \epsilon \sim \mathcal{N}(0, \sigma = 2).
\end{equation}

The results for the heterocestic case are included in Fig.\ref{fig:results}b following the same layout as in the bimodal case. Again, the exact GP is unable to produce heteroscedastic predictive distributions. On the other hand, the variance in SIP's predictive distribution clearly depends on the location in the $x$-axis for the predictions, following the same heteroscedastic pattern present in the original data. In this case, we can also see that the sampled functions from the prior GP in SIP are much smoother, pointing to a different fit for the hyperparameters of the GP prior employed.

Overall, we see that the predictive distribution of SIP is far more representative of the original data than the resulting distribution from the GP. These results are also supported by the error metrics in Table \ref{tab:results}. Here, the RMSE for both methods are almost equivalent, especially when taking into account their respective standard errors. This can be understood directly as both of their predictive means lay in the same vicinity. However, in terms of the negative log-likelihood and the \textit{continous ranked probability score} \cite{gneiting2007strictly} SIP is clearly superior. These results can be explained by the fact that the predictive mean of both methods almost coincide due to the data employed, while the predictive distribution of SIP is much closer to the original data than the one of the GP. This suggests that SIP successfully avoids the overly-simplistic Gaussian prediction bias imposed by the GP prior, which would also result in Gaussian predictions. Therefore, through its more general and flexible formulation, SIP is capable of bypassing formulation biases that may hinder the predictions and captures otherwise neglected key features present in the data.

\section{Conclusions}
\label{sec:conclusions}

We have shown that SIP avoids formulation bias whenever the data strongly deviates from the restrictions imposed by the selected model. We extend the results in the original contribution \cite{santana2022sparse} by using a GP prior, making the exact posterior available, which is also Gaussian. However, if the data differs from the expected exact GP predictions, SIP's framework provides enough flexibility so that its predictive distribution can turn out not necessarily Gaussian, even when using the aforementioned GP prior, which is a wrong assumption. In fact, SIP's predictive distribution can be much more representative of the original data, reflecting complex features that would be otherwise neglected, \emph{e.g.} bimodality and heteroscedasticity. This complements the results in the original paper, where it is also shown that HMC is not capable of capturing these complex patterns in the data. Further work will include studying the robustness of function-space-based methods to adversarial examples and a more detailed analysis of the inference process conducted \cite{knoblauch2019generalized}, as well as applications of SIP to complex real-world situations.

\section*{Acknowledgements}

Authors gratefully acknowledge the use of the facilities of Centro de Computacion Cientifica (CCC) at Universidad Autónoma de Madrid. The authors also acknowledge financial support from Spanish Plan Nacional I+D+i, PID2019-106827GB-I00. SRS acknowledges the BBVA Foundation project and the Trustonomy project, which have received funding from the European Community’s Horizon 2020 research and innovation programme under grant agreement No 812003. BZ has been supported by the Programa Atracci\'on de Talento de la Comunidad de Madrid under grant n. 2017- T2/TIC-5455, from the Comunidad de Madrid/UAM “Proyecto de J\'ovenes Investigadores” grant n. SI1/PJI/2019-00294, as well as from Spanish “Proyectos de I+D de Generaci\'on de Conocimiento” via grants PGC2018-096646-A-I00 and PGC2018-095161-B-I00. BZ finally acknowledge the support from Generalitat Valenciana through the plan GenT program (CIDEGENT/2020/055).

\bibliography{bib_SIPs}
\bibliographystyle{wfvml2022}

\end{document}